\documentclass[11pt,a4paper]{article}
\usepackage[hyperref]{acl2018}
\usepackage{times}
\usepackage{latexsym}
\usepackage{amsmath}
\DeclareMathOperator{\sign}{sign}
\usepackage{amssymb,amsmath,amsthm,enumitem}
\usepackage{comment}

\usepackage{url}
\usepackage{graphics}

\usepackage{times}
\usepackage{graphicx}
\usepackage{amsmath,amssymb,enumerate,bbm,color,alltt}
\usepackage{graphicx}
\usepackage{float}
\usepackage{stfloats}
\usepackage{subfig}
\usepackage{longtable}
\usepackage{tabularx}
\usepackage{multirow}
\usepackage{booktabs}
\usepackage{subfig}

\aclfinalcopy 
\def\aclpaperid{1022} 

\title{NASH: Toward End-to-End Neural Architecture for Generative Semantic Hashing}

\author{Dinghan Shen$^{\mathbf{1}*}$, ~Qinliang Su$^{\mathbf{2}}\thanks{~~~Equal contribution.}~~$, ~Paidamoyo Chapfuwa$^{\mathbf{1}}$,
	\smallskip 
	~  \\
	\bf{~	Wenlin Wang$^{\mathbf{1}}$, Guoyin Wang$^{\mathbf{1}}$, ~Lawrence Carin$^{\mathbf{1}}$}, ~Ricardo Henao$^{\mathbf{1}}$ \\
	\smallskip 
	$^{\mathbf{1}}$ Duke University~~~~~~~~~~~~ 
	$^{\mathbf{2}}$ Sun Yat-sen University \\
	{\tt dinghan.shen@duke.edu} 
	}

\date{}

\begin{document}
\maketitle
\begin{abstract}
Semantic hashing has become a powerful paradigm for fast similarity search in many information retrieval systems.
While fairly successful, previous techniques generally require two-stage training, and the binary constraints are handled \emph{ad-hoc}.
In this paper, we present an \emph{end-to-end} Neural Architecture for Semantic Hashing (NASH), where the binary hashing codes are treated as \emph{Bernoulli} latent variables.
A neural variational inference framework is proposed for training, where gradients are directly backpropagated through the discrete latent variable to optimize the hash function.
We also draw connections between proposed method and \emph{rate-distortion theory}, which provides a theoretical foundation for the effectiveness of the proposed framework.
Experimental results on three public datasets demonstrate that our method significantly outperforms several state-of-the-art models on both \emph{unsupervised} and \emph{supervised} scenarios.
\end{abstract}
\section{Introduction}
The problem of \emph{similarity search}, also called \emph{nearest-neighbor search}, consists of finding documents from a large collection of documents, or \emph{corpus}, which are most similar to a query document of interest.
Fast and accurate similarity search is at the core of many information retrieval applications, such as plagiarism analysis \citep{stein2007strategies}, collaborative filtering \citep{koren2008factorization}, content-based multimedia retrieval \citep{lew2006content} and caching \citep{pandey2009nearest}.
Semantic hashing is an effective approach for fast similarity search \citep{salakhutdinov2009semantic, zhang2010self, wang2014hashing}.
By representing every document in the corpus as a similarity-preserving discrete (binary) \emph{hashing code}, the similarity between two documents can be evaluated by simply calculating pairwise Hamming distances between hashing codes, \emph{i.e.}, the number of bits that are different between two codes.
Given that today, an ordinary PC is able to execute millions of Hamming distance computations in just a few milliseconds \cite{zhang2010self}, this semantic hashing strategy is very computationally attractive.

While considerable research has been devoted to text (semantic) hashing, existing approaches typically require two-stage training procedures.
These methods can be generally divided into two categories: 
($i$) binary codes for documents are first learned in an unsupervised manner, then $l$ binary classifiers are trained via supervised learning to predict the $l$-bit hashing code \cite{zhang2010self, xu2015convolutional}; 
($ii$) continuous text representations are first inferred, which are binarized as a second (separate) step during testing \cite{wang2013semantic, chaidaroon2017variational}.
Because the model parameters are not learned in an end-to-end manner,
these two-stage training strategies may result in suboptimal local optima.
This happens because different modules within the model are optimized separately, preventing the sharing of information between them.
Further, in existing methods, binary constraints are typically handled \emph{ad-hoc} by truncation, \emph{i.e.}, the hashing codes are obtained via direct binarization from continuous representations after training.
As a result, the information contained in the continuous representations is lost during the (separate) binarization process.
Moreover, training different modules (mapping and classifier/binarization) separately often
requires additional hyperparameter tuning for each training stage, which can be laborious and time-consuming.

In this paper, we propose a simple and generic neural architecture for text hashing that learns binary latent codes for documents in an \emph{end-to-end} manner.
Inspired by recent advances in neural variational inference (NVI) for text processing  \citep{miao2016neural, yang2017improved, shen2017deconvolutional}, we approach semantic hashing from a generative model perspective, where binary (hashing) codes are represented as either \emph{deterministic} or \emph{stochastic} Bernoulli latent variables.
The inference (encoder) and generative (decoder) networks are optimized jointly by maximizing a variational lower bound to the marginal distribution of input documents (corpus).
By leveraging a simple and effective method to estimate the gradients with respect to discrete (binary) variables, the loss term from the generative (decoder) network can be directly backpropagated into the inference (encoder) network to optimize the hash function.

Motivated by the \emph{rate-distortion theory} \citep{berger1971rate, theis2017lossy}, we propose to inject data-dependent noise into the latent codes during the decoding stage, which adaptively accounts for the tradeoff between minimizing \emph{rate} (number of bits used, or effective code length) and \emph{distortion} (reconstruction error) during training.
The connection between the proposed method and \emph{rate-distortion theory} is further elucidated, providing a theoretical foundation for the effectiveness of our framework.

Summarizing, the contributions of this paper are: (\emph{\romannumeral1}) to the best of our knowledge, we present the first semantic hashing architecture that can be trained in an \emph{end-to-end} manner;
(\emph{\romannumeral2}) we propose a \emph{neural variational inference} framework to learn compact (regularized) binary codes for documents, achieving promising results on both \emph{unsupervised} and \emph{supervised} text hashing; (\emph{\romannumeral3}) the connection between our method and \emph{rate-distortion theory} is established, from which we demonstrate the advantage of injecting \emph{data-dependent noise} into the latent variable during training.

\section{Related Work}
Models with discrete random variables have attracted much attention in the deep learning community \citep{jang2016categorical, maddison2016concrete, van2017neural, li2017deep, shu2017compressing}.
Some of these structures are more natural choices for language or speech data, which are inherently discrete.
More specifically, \citet{van2017neural} combined VAEs with vector quantization to learn discrete latent representation, and demonstrated the utility of these learned representations on images, videos, and speech data.
\citet{li2017deep} leveraged both pairwise label and classification information to learn discrete hash codes, which exhibit state-of-the-art performance on image retrieval tasks.

For natural language processing (NLP), although significant research has been made to learn \emph{continuous} deep representations for words or documents \cite{mikolov2013distributed, kiros2015skip, Shen2018Baseline}, \emph{discrete} neural representations have been mainly explored in learning word embeddings \cite{shu2017compressing, chen2017learning}. In these recent works, words are represented as a vector of discrete numbers, which are very efficient storage-wise, while showing comparable performance on several NLP tasks, relative to continuous word embeddings.
However, discrete representations that are learned in an \emph{end-to-end} manner at the \emph{sentence} or \emph{document} level have been rarely explored.
Also there is a lack of strict evaluation regarding their effectiveness.
Our work focuses on learning discrete (binary) representations for text documents.
Further, we employ semantic hashing (fast similarity search) as a mechanism to evaluate the quality of learned binary latent codes.

\begin{figure}[!t]
	\centering
	\vspace{-2mm}
	\includegraphics[width=.4\textwidth]{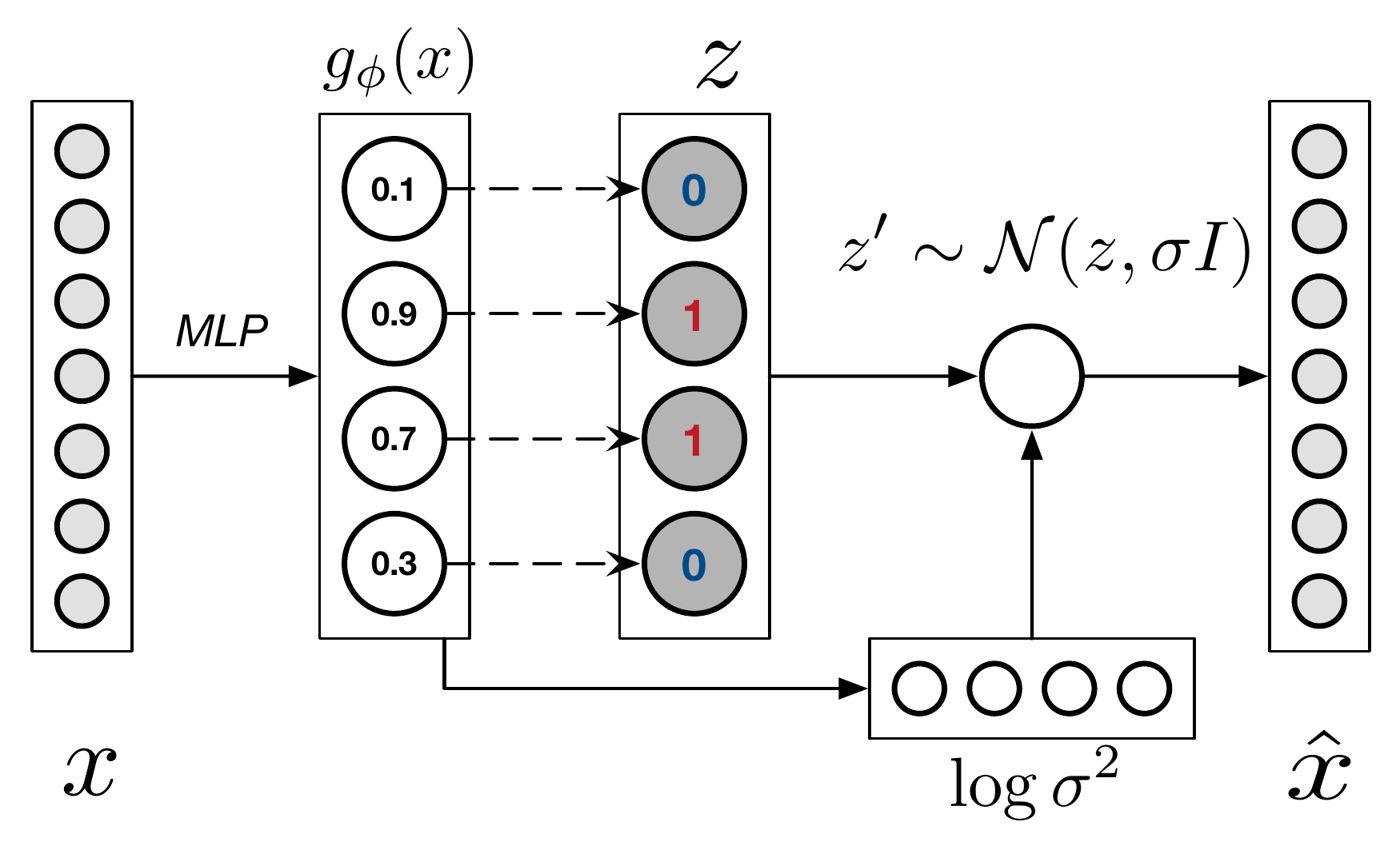}
	\caption{NASH for \emph{end-to-end} semantic hashing. The inference network maps $x\to z$ using an MLP and the generative network recovers $x$ as $z\to \hat{x}$.}
	\label{fig:model}
	\vspace{-2mm}
\end{figure}


\section{The Proposed Method}\label{sec:method}
%

\subsection{Hashing under the NVI Framework}
Inspired by the recent success of variational autoencoders for various NLP problems \citep{miao2016neural, bowman2015generating, yang2017improved, miao2017discovering, shen2017deconvolutional, wang2018topic}, we approach the training of discrete (binary) latent variables from a generative perspective.
Let $x$ and $z$ denote the input document and its corresponding binary hash code, respectively.
Most of the previous text hashing methods focus on modeling the encoding distribution $p(z|x)$, or \emph{hash function}, so the local/global pairwise similarity structure of documents in the original space is preserved in latent space \citep{zhang2010self, wang2013semantic, xu2015convolutional, wang2014hashing}.
However, the generative (decoding) process of reconstructing $x$ from binary latent code $z$, \emph{i.e.}, modeling distribution $p(x|z)$, has been rarely considered.
Intuitively, latent codes learned from a model that accounts for the generative term should naturally encapsulate key semantic information from $x$ because the generation/reconstruction objective is a function of $p(x|z)$.
In this regard, the generative term provides a natural training objective for semantic hashing.

We define a generative model that simultaneously accounts for both the encoding distribution, $p(z|x)$, and decoding distribution, $p(x|z)$, by defining approximations $q_\phi(z|x)$ and $q_\theta(x|z)$, via inference and generative networks, $g_\phi(x)$ and $g_\theta(z)$, parameterized by $\phi$ and $\theta$, respectively.
Specifically, $x \in\mathcal{Z}_+^{|V|}$ is the bag-of-words (count) representation for the input document, where $|V|$ is the vocabulary size.
Notably, we can also employ other count weighting schemes as input features $x$, \emph{e.g.}, the term frequency-inverse document frequency (TFIDF) \cite{manning2008introduction}.
For the encoding distribution, a latent variable $z$ is first inferred from the input text $x$, by constructing an inference network $g_{\phi}(x)$ to approximate the true posterior distribution $p(z|x)$ as $q_\phi(z|x)$.
Subsequently, the decoder network $g_\theta (z)$ maps $z$ back into input space to reconstruct the original sequence $x$ as $\hat{x}$, approximating $p(x|z)$ as $q_\theta(x|z)$ (as shown in Figure~\ref{fig:model}).
This \emph{cyclic} strategy, $x\to z \to \hat{x}\approx x$, provides the latent variable $z$ with a better ability to generalize \cite{miao2016neural}.

To tailor the NVI framework for semantic hashing, we cast $z$ as a binary latent variable and assume a multivariate Bernoulli prior on $z$: $p(z) \sim \text{Bernoulli}(\gamma) = \prod_{i=1}^l \gamma_i^{z_i}(1 - \gamma_i)^{1-z_i}$, where $\gamma_i \in [0, 1]$ is component $i$ of vector $\gamma$.
Thus, the encoding (approximate posterior) distribution $q_\phi(z|x)$ is restricted to take the form  $q_\phi(z|x) = \text{Bernoulli}(h)$, where $h=\sigma(g_\phi(x))$, $\sigma(\cdot)$ is the sigmoid function, and $g_\phi(\cdot)$ is the (nonlinear) inference network specified as a multilayer perceptron (MLP).
As illustrated in Figure~\ref{fig:model}, we can obtain samples from the Bernoulli posterior either \emph{deterministically} or \emph{stochastically}.
Suppose $z$ is a $l$-bit hash code, for the \emph{deterministic} binarization, we have, for $i = 1,2,......,l$:
\begin{align}
z_i = & \ \boldsymbol{1}_{\sigma(g^i_\phi(x)) > 0.5} = & \ \frac{{\sign}(\sigma(g^i_\phi(x)-0.5) + 1}{2}, \label{eq:deter}
\end{align}
where $z$ is the binarized variable, and $z_i$ and $g^i_\phi(x)$ denote the $i$-th dimension of $z$ and $g_\phi(x)$, respectively.
The standard Bernoulli sampling in \eqref{eq:deter} can be understood as setting a hard threshold at 0.5 for each representation dimension, therefore, the binary latent code is generated deterministically.
Another strategy to obtain the discrete variable is to binarize $h$ in a \emph{stochastic} manner:
\begin{align}
z_i = & \ \boldsymbol{1}_{\sigma(g^i_\phi(x)) > \mu_i} = & \ \frac{{\sign}(\sigma(g^i_\phi(x)) -\mu_i) + 1}{2}, \label{eq:stoc}
\end{align}
where $\mu_i\sim{\rm Uniform}(0, 1)$.
Because of this sampling process, we do not have to assume a pre-defined threshold value like in \eqref{eq:deter}.

\subsection{Training with Binary Latent Variables}
To estimate the parameters of the encoder and decoder networks, we would ideally maximize the marginal distribution $p(x) = \int p(z)p(x|z)dz$.
However, computing this marginal is intractable in most cases of interest.
Instead, we maximize a variational lower bound.
This approach is typically employed in the VAE framework \cite{kingma2013auto}:
\begin{align}
\vspace{-3mm}
& \mathcal{L}_{\rm vae} = \mathbb{E}_{q_\phi(z|x)} \left[\log \frac{q_\theta(x|z)p(z)}{q_\phi(z|x)}\right], \label{eq:vae} \\
& = \mathbb{E}_{q_\phi(z|x)}[\log q_\theta(x|z)] - D_{KL}(q_\phi(z|x)||p(z)) , \notag
\vspace{-3mm}
\end{align}
where the Kullback-Leibler (KL) divergence $D_{KL}(q_\phi(z|x)||p(z))$ encourages the approximate posterior distribution $q_\phi(z|x)$ to be close to the multivariate Bernoulli prior $p(z)$.
In this case, $D_{KL}(q_\phi(z|x)|p(z))$ can be written in closed-form as a function of $g_\phi(x)$:
\begin{align}
\vspace{-3mm}
D_{KL} = & \ g_\phi(x)\log \frac{g_\phi(x)}{\gamma} \nonumber \\
+ & \ (1 - g_\phi(x)) \log \frac{1 - g_\phi(x)}{1 - \gamma} .
 \label{eq:kl}
 \vspace{-3mm}
\end{align}
Note that the gradient for the KL divergence term above can be evaluated easily.

For the first term in \eqref{eq:vae}, we should in principle estimate the influence of $\mu_i$ in \eqref{eq:stoc} on $q_\theta(x|z)$ by averaging over the entire (uniform) noise distribution.
However, a closed-form distribution does not exist since it is not possible to enumerate all possible configurations of $z$, especially when the latent dimension is large.
Moreover, discrete latent variables are inherently incompatible with backpropagation, since the derivative of the sign function is zero for almost all input values.
As a result, the exact gradients of $L_{\rm vae}$ wrt the inputs before binarization would be essentially all zero.

To estimate the gradients for binary latent variables, we utilize the straight-through (ST) estimator, which was first introduced by \citet{hinton2012neural}. 
So motivated, the strategy here is to simply backpropagate through the hard threshold by approximating the gradient $\partial z / \partial \phi $ as 1.
Thus, we have:
\begin{align}
\vspace{-3mm}
&\frac{d \mathbb{E}_{q_\phi(z|x)}[\log q_\theta(x|z)] }{\partial \phi}  \nonumber \\
&\quad = \frac{d \mathbb{E}_{q_\phi(z|x)}[\log q_\theta(x|z)] }{d z} \frac{d z}{d \sigma(g^i_\phi(x))} \frac{d \sigma(g^i_\phi(x))}{d \phi} \nonumber \\
&\quad \approx \frac{d \mathbb{E}_{q_\phi(z|x)}[\log q_\theta(x|z)] }{d z}  \frac{d \sigma(g^i_\phi(x))}{d \phi}
\vspace{-3mm}
\end{align}
Although this is clearly a biased estimator, it has been  shown to be a fast and efficient method relative to other gradient estimators for discrete variables, especially for the Bernoulli case \citep{bengio2013estimating, hubara2016binarized, theis2017lossy}.
With the ST gradient estimator, the first loss term in \eqref{eq:vae} can be backpropagated into the encoder network to fine-tune the hash function $g_\phi(x)$.

For the approximate generator $q_\theta(x|z)$ in \eqref{eq:vae}, let $x_i$ denote the one-hot representation of $i$th word within a document.
Note that $x=\sum_i x_i$ is thus the bag-of-words representation for document $x$.
To reconstruct the input $x$ from $z$, we utilize a \emph{softmax} decoding function written as:
\begin{align}
\vspace{-3mm}
q(x_i=w|z) & = \frac{\exp(z^T E x_w + b_w)}{\sum_{j = 1}^{|V|} \exp(z^T E x_j + b_j)} ,
\label{eq:decoder}
\vspace{-3mm}
\end{align}
where $q(x_i=w|z)$ is the probability that $x_i$ is word $w\in V$, $q_\theta(x|z)=\prod_i q(x_i=w|z)$ and $\theta=\{E,b_1,\ldots,b_{|V|}\}$.
Note that $E \in \mathbb{R}^{d \times |V|}$ can be interpreted as a word embedding matrix to be learned, and $\{b_i\}_{i=1}^{|V|}$ denote bias terms.
Intuitively, the objective in \eqref{eq:decoder} encourages the discrete vector $z$ to be close to the embeddings for every word that appear in the input document $x$.
As shown in Section~\ref{embeddings}, meaningful semantic structures can be learned and manifested in the word embedding matrix $E$.

\subsection{Injecting Data-dependent Noise to $z$}
To reconstruct text data $x$ from sampled binary representation $z$, a deterministic decoder is typically utilized \cite{miao2016neural, chaidaroon2017variational}.
Inspired by the success of employing stochastic decoders in image hashing applications \cite{dai2017stochastic, theis2017lossy}, in our experiments, we found that injecting random Gaussian noise into $z$ makes the decoder a more favorable regularizer for the binary codes, which in practice leads to stronger retrieval performance.
Below, we invoke the \emph{rate-distortion theory} to perform some further analysis, which leads to interesting findings.

Learning binary latent codes $z$ to represent a continuous distribution $p(x)$ is a classical information theory concept known as \emph{lossy source coding}.
From this perspective, semantic hashing, which compresses an input document into compact binary codes, can be casted as a conventional \emph{rate-distortion tradeoff} problem \cite{theis2017lossy, balle2016end}:
\begin{align}
\vspace{-8mm}
\min \ \underbrace{- \log _{2} R(z)}_{\text{Rate}} + \beta \underbrace{ \cdot D (x, \hat{x})}_{\text{Distortion}} \,,
\label{eq:rate_dist}
\vspace{-5mm}
\end{align}
where \emph{rate} and \emph{distortion} denote the effective code length, \emph{i.e.}, the number of bits used, and the distortion introduced by the encoding/decoding sequence, respectively.
Further, $\hat{x}$ is the reconstructed input and $\beta$ is a hyperparameter that controls the tradeoff between the two terms.

Considering the case where we have a Bernoulli \emph{prior} on $z$ as $p(z) \sim \text{Bernoulli}(\gamma)$, and $x$ conditionally drawn from a Gaussian distribution $p(x|z)  \sim \mathcal{N} (Ez, \sigma^2I)$.
Here, $E = \{e_i\}_{i=1}^{|V|}$, where $e_i \in \mathbb{R}^d$ can be interpreted as a \emph{codebook} with $|V|$ \emph{codewords}.
In our case, $E$ corresponds to the \emph{word embedding matrix} as in \eqref{eq:decoder}.

For the case of stochastic latent variable $z$, the objective function in \eqref{eq:vae} can be written in a form similar to the \emph{rate-distortion} tradeoff:
\begin{align}
\small {
	\vspace{-5mm}
	\min  \mathbb{E}_{q_\phi(z|x)}\left[ \underbrace{-\log q_{\phi}(z|x)}_{\text{Rate}}  +\underbrace{ \frac{1}{2\sigma^2}}_{\beta}  \underbrace{|| x - Ez||^2_2}_{\text{Distortion}}+ C\right]\,,
	\vspace{-5mm}
}
\end{align}
where $C$ is a constant that encapsulates the prior distribution $p(z)$ and the Gaussian distribution normalization term.
Notably, the trade-off hyperparameter $\beta = \sigma^{-2}/2$ is closely related to the variance of the distribution $p(x|z)$.
In other words, by controlling the variance $\sigma$, the model can adaptively explore different trade-offs between the \emph{rate} and \emph{distortion} objectives. However, the optimal trade-offs for distinct samples may  be different.

Inspired by the observations above, we propose to inject data-dependent noise into latent variable $z$, rather than to setting the variance term $\sigma^2$ to a fixed value \cite{dai2017stochastic, theis2017lossy}. 
Specifically, $\log\sigma^2$ is obtained via a one-layer MLP transformation from $g_\phi(x)$. Afterwards, we sample $z^{\prime}$ from $\mathcal{N} (z, \sigma^2I)$, which then replace $z$ in \eqref{eq:decoder} to infer the probability of generating individual words (as shown in Figure~\ref{fig:model}).
As a result, the variances are different for every input document $x$, and thus the model is provided with additional flexibility to explore various trade-offs between \emph{rate} and \emph{distortion} for different training observations.
Although our decoder is not a strictly Gaussian distribution, as in \eqref{eq:decoder}, we found empirically that injecting data-dependent noise into $z$ yields strong retrieval results, see Section~\ref{hashing}.

\subsection{Supervised Hashing}
The proposed Neural Architecture for Semantic Hashing (NASH) can be extended to supervised hashing, where a mapping from latent variable $z$ to labels $y$ is learned, here parametrized by a two-layer MLP followed by a fully-connected softmax layer.
To allow the model to explore and balance between maximizing the variational lower bound in \eqref{eq:vae} and minimizing the discriminative loss, the following joint training objective is employed:
\begin{align}
\mathcal{L} & = - \mathcal{L}_{\rm vae}(\theta, \phi; x) + \alpha \mathcal{L}_{\rm dis}(\eta; z, y).
\label{eq:sup_loss}
\end{align} 
where $\eta$ refers to parameters of the MLP classifier and $\alpha$ controls the relative weight between the variational lower bound ($\mathcal{L}_{\rm vae}$) and discriminative loss ($\mathcal{L}_{\rm dis}$), defined as the cross-entropy loss.
The parameters $\{\theta, \phi, \eta\}$ are learned end-to-end \emph{via} Monte Carlo estimation.

\section{Experimental Setup}
\subsection{Datasets}
We use the following three standard publicly available datasets for training and evaluation: ($i$) \emph{Reuters21578}, containing 10,788 news documents, which have been classified into 90 different categories.
($ii$) \emph{20Newsgroups}, a collection of 18,828 newsgroup documents, which are categorized into 20 different topics.
($iii$) TMC (stands for SIAM text mining competition), containing air
traffic reports provided by NASA.
TMC consists 21,519 training documents divided into 22 different categories.
To make direct comparison with prior works, we employed the TFIDF features on these datasets supplied by \cite{chaidaroon2017variational}, where the vocabulary sizes for the three datasets are set to 10,000, 7,164 and 20,000, respectively.

\subsection{Training Details}
For the inference networks, we employ a feed-forward neural network with 2 hidden layers (both with 500 units) using the ReLU non-linearity activation function, which transform the input documents, \emph{i.e.}, TFIDF features in our experiments, into a continuous representation.
Empirically, we found that stochastic binarization as in \eqref{eq:stoc} shows stronger performance than deterministic binarization, and thus use the former in our experiments.
However, we further conduct a systematic ablation study in Section~\ref{sec:ablation} to compare the two binarization strategies.

Our model is trained using Adam \cite{kingma2014adam}, with a learning rate of $1 \times 10^{-3}$ for all parameters.
We decay the learning rate by a factor of 0.96 for every 10,000 iterations.
Dropout \cite{srivastava2014dropout} is employed on the output of encoder networks, with the rate selected from $\{0.7, 0.8, 0.9\}$ on the validation set.
To facilitate comparisons with previous methods, we set the dimension of $z$, \emph{i.e.}, the number of bits within the hashing code) as 8, 16, 32, 64, or 128.

\subsection{Baselines}
We evaluate the effectiveness of our framework on both unsupervised and supervised semantic hashing tasks.
We consider the following \emph{unsupervised} baselines for comparisons: Locality Sensitive Hashing (LSH) \citep{datar2004locality}, Stack Restricted Boltzmann Machines (S-RBM) \citep{salakhutdinov2009semantic}, Spectral Hashing (SpH) \citep{weiss2009spectral}, Self-taught Hashing (STH) \citep{zhang2010self} and Variational Deep Semantic Hashing (VDSH) \citep{chaidaroon2017variational}. 

For supervised semantic hashing, we also compare NASH against a number of baselines: Supervised Hashing with Kernels (KSH) \citep{liu2012supervised}, Semantic Hashing using Tags and Topic Modeling (SHTTM) \citep{wang2013semantic} and Supervised VDSH \citep{chaidaroon2017variational}. 
It is worth noting that unlike all these baselines, our NASH model is trained end-to-end in one-step.

\subsection{Evaluation Metrics}
To evaluate the hashing codes for similarity search, we consider each document in the testing set as a query document.
Similar documents to the query in the corresponding training set need to be retrieved based on the Hamming distance of their hashing codes, \emph{i.e.} number of different bits.
To facilitate comparison with prior work \citep{wang2013semantic, chaidaroon2017variational}, the performance is measured with precision. 
Specifically, during testing, for a query document, we first retrieve the 100 nearest/closest documents according to the Hamming distances of the corresponding hash codes (i.e., the number of different bits). 
We then examine the percentage of documents among these 100 retrieved ones that belong to the same label (topic) with the query document (we consider documents having the same label as relevant pairs).
 The ratio of the number of relevant documents to the number of retrieved documents (fixed value of 100) is calculated as the precision score. 
 The precision scores are further averaged over all test (query) documents. 


\section{Experimental Results} 

\begin{table}
	\centering
	\resizebox{\columnwidth}{!}{%
		\begin{tabular}{c||c|c|c|c|c}
			\toprule[1.2pt]
			\textbf{Method} &  	\textbf{8 bits} & \textbf{16 bits} & 	\textbf{32 bits} & \textbf{64 bits} & \textbf{128 bits} \\
			\hline
			LSH        & 0.2802 & 0.3215 & 0.3862 & 0.4667  & 0.5194 \\
			S-RBM    & 0.5113 & 0.5740 & 0.6154 & 0.6177 & 0.6452  \\ 
			SpH        & 0.6080 & 0.6340 & 0.6513 & 0.6290 & 0.6045   \\
			STH        & 0.6616 & 0.7351 & 0.7554 & 0.7350 & 0.6986  \\ 
			VDSH           & 0.6859 & 0.7165 & 0.7753 & 0.7456 & 0.7318  \\ 
			\hline
			NASH & 0.7113 & 0.7624 & 0.7993 & 0.7812  & 0.7559 \\
			NASH-N  & 0.7352 & 0.7904 & 0.8297 & 0.8086  & 0.7867 \\
			NASH-DN   & \textbf{0.7470} & \textbf{0.8013}  & \textbf{0.8418}  & \textbf{0.8297}  & \textbf{0.7924}  \\
			\bottomrule[1.2pt]
		\end{tabular}
	}
	\caption{Precision of the top 100 retrieved documents on \emph{Reuters} dataset (\emph{Unsupervised hashing}).}
	\label{tab:reu}
	\vspace{-1mm}
\end{table}

\begin{figure}[t!]
	\centering
	\includegraphics[width=.4\textwidth]{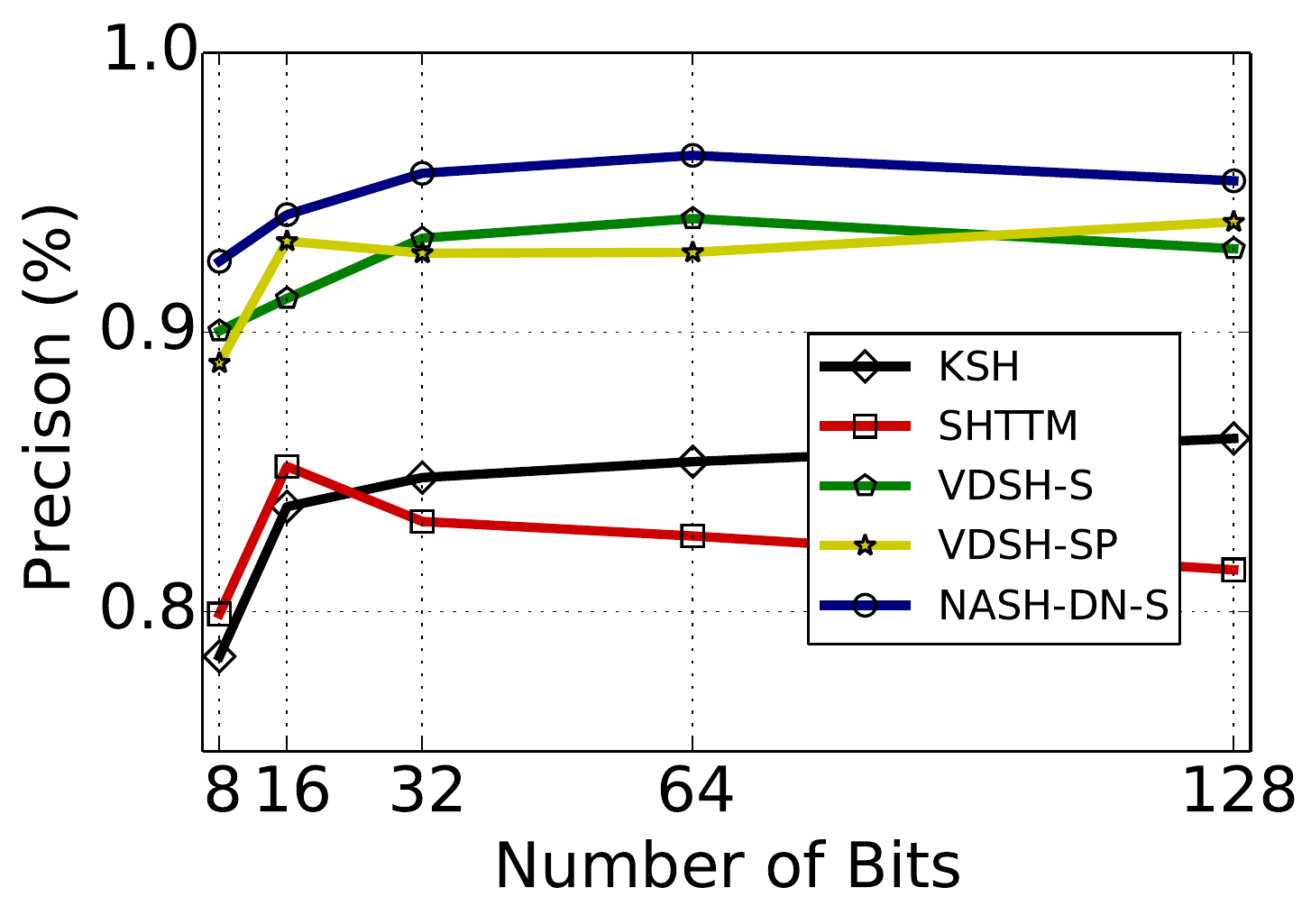}
	\caption{Precision of the top 100 retrieved documents on \emph{Reuters} dataset (\emph{Supervised hashing}), compared with other supervised baselines. }
	\vspace{-1mm}
	\label{fig:sup}
\end{figure}

We experimented with four variants for our NASH model: (\emph{\romannumeral1}) NASH: with deterministic decoder; (\emph{\romannumeral2}) NASH-N: with \emph{fixed} random noise injected to decoder; (\emph{\romannumeral3}) NASH-DN: with \emph{data-dependent} noise injected to decoder; (\emph{\romannumeral4}) NASH-DN-S: NASH-DN with supervised information during training. 

\subsection{Semantic Hashing Evaluation} \label{hashing}

Table~\ref{tab:reu} presents the results of all models on Reuters dataset.
Regarding unsupervised semantic hashing, all the NASH variants consistently outperform the baseline methods by a substantial margin, indicating that our model makes the most effective use of unlabeled data and manage to assign similar hashing codes, \emph{i.e.}, with small Hamming distance to each other, to documents that belong to the same label.
It can be also observed that the injection of noise into the decoder networks has improved the robustness of learned binary representations, resulting in better retrieval performance.
More importantly, by making the variances of noise adaptive to the specific input, our NASH-DN achieves even better results, compared with NASH-N, highlighting the importance of exploring/learning the trade-off between rate and distortion objectives by the data itself.
We observe the same trend and superiority of our NASH-DN models on the other two benchmarks, as shown in Tables~\ref{tab:news} and \ref{tab:tmc}. 

\begin{table*}[ht!]  \small
	\centering
	\begin{tabular}{c||c|c|c|c|c|c} 
		\toprule[1.2pt]
		Word &  	\textbf{weapons} & \textbf{medical} & 	\textbf{companies} & \textbf{define} & \textbf{israel} & \textbf{book}  \\
		\hline
		\multirow{5}{*}{NASH}& gun & treatment & company & definition & israeli & books\\
		& guns  & disease & market & defined & arabs & english \\
		& weapon & drugs & afford & explained & arab & references \\
		& armed & health & products & discussion & jewish & learning \\
		& assault  & medicine & money & knowledge & jews & reference \\
		\hline
		\multirow{5}{*}{NVDM}& guns & medicine & expensive & defined & israeli & books\\
		& weapon  & health & industry & definition & arab & reference \\
		& gun & treatment & company & printf & arabs & guide \\
		& militia & disease & market & int & lebanon & writing \\
		& armed  & patients & buy & sufficient & lebanese & pages \\
		\bottomrule[1.2pt]
	\end{tabular}
	\caption{The five nearest words in the semantic space learned by NASH, compared with the results from NVDM \cite{miao2016neural}.}
	\label{tab:words}
	\vspace{-1mm}
\end{table*}

\begin{table} [!h]
	\centering
	\resizebox{\columnwidth}{!}{%
		\begin{tabular}{c||c|c|c|c|c}
			\toprule[1.2pt]
			\textbf{Method} &  	\textbf{8 bits} & \textbf{16 bits} & 	\textbf{32 bits} & \textbf{64 bits} & \textbf{128 bits} \\
			\hline
			\multicolumn{6}{c}{\emph{Unsupervised Hashing}} \\
			\hline
			LSH        & 0.0578 & 0.0597 & 0.0666  & 0.0770  & 0.0949  \\
			S-RBM    & 0.0594 & 0.0604 & 0.0533 & 0.0623 & 0.0642   \\ 
			SpH        & 0.2545 & 0.3200 & 0.3709 & 0.3196 & 0.2716   \\
			STH        & 0.3664 & 0.5237 & 0.5860 & \textbf{0.5806} & \textbf{0.5443}  \\ 
			VDSH           & 0.3643 & 0.3904 & 0.4327 & 0.1731 & 0.0522  \\ 
			\hline
			NASH & 0.3786 & 0.5108 & 0.5671 & 0.5071  & 0.4664  \\
			NASH-N  & 0.3903  & 0.5213 & 0.5987  & 0.5143  & 0.4776  \\
			NASH-DN   & \textbf{0.4040} & \textbf{0.5310}  & \textbf{0.6225}  & 0.5377  & 0.4945  \\
			\hline
			\multicolumn{6}{c}{\emph{Supervised Hashing}} \\
			\hline
			KSH & 0.4257 & 0.5559 & 0.6103 & 0.6488 & 0.6638 \\
			SHTTM  & 0.2690 & 0.3235 & 0.2357 & 0.1411 & 0.1299 \\
			VDSH-S & 0.6586 & 0.6791 & 0.7564&  0.6850& 0.6916 \\
			VDSH-SP & \textbf{0.6609} & 0.6551 & 0.7125 & 0.7045 & 0.7117  \\
			NASH-DN-S   & 0.6247 & \textbf{0.6973}  & \textbf{0.8069}  & \textbf{0.8213}  & \textbf{0.7840}  \\
			\bottomrule[1.2pt]
		\end{tabular}
	}
	\caption{Precision of the top 100 retrieved documents on \emph{20Newsgroups} dataset. }
	\label{tab:news}
	\vspace{-1mm}
\end{table}
 
 \begin{table} [ht!]
 	\centering
 	\resizebox{\columnwidth}{!}{%
 		\begin{tabular}{c||c|c|c|c|c}
 			\toprule[1.2pt]
 			\textbf{Method} &  	\textbf{8 bits} & \textbf{16 bits} & 	\textbf{32 bits} & \textbf{64 bits} & \textbf{128 bits} \\
 			\hline
 			\multicolumn{6}{c}{\emph{Unsupervised Hashing}} \\
 			\hline
 			LSH        & 0.4388 & 0.4393 & 0.4514 & 0.4553 & 0.4773 \\
 			S-RBM    & 0.4846 & 0.5108 & 0.5166 & 0.5190 & 0.5137  \\ 
 			SpH        & 0.5807 & 0.6055 & 0.6281 & 0.6143 & 0.5891   \\
 			STH        & 0.3723 & 0.3947 & 0.4105 & 0.4181 & 0.4123  \\ 
 			VDSH           & 0.4330 & 0.6853 & 0.7108 & 0.4410 & 0.5847  \\ 
 			\hline
 			NASH & 0.5849 & 0.6573 & 0.6921 & 0.6548 & 0.5998 \\
 			NASH-N  & 0.6233 & 0.6759 & 0.7201 & 0.6877 & 0.6314 \\
 			NASH-DN   & \textbf{0.6358} & \textbf{0.6956}  & \textbf{0.7327}  & \textbf{0.7010}  & \textbf{0.6325}  \\
 			\hline
 			\multicolumn{6}{c}{\emph{Supervised Hashing}} \\
 			\hline
 			KSH & 0.6608 & 0.6842 & 0.7047 & 0.7175 & 0.7243 \\
 			SHTTM  & 0.6299 & 0.6571 & 0.6485 & 0.6893 & 0.6474 \\
 			VDSH-S & 0.7387 & 0.7887 & 0.7883 & 0.7967 & 0.8018  \\
 			VDSH-SP & \textbf{0.7498} & 0.7798 & 0.7891 & 0.7888 & 0.7970 \\
 			NASH-DN-S   & 0.7438 & \textbf{0.7946} & \textbf{0.7987} & \textbf{0.8014} & \textbf{0.8139}    \\
 			\bottomrule[1.2pt]
 		\end{tabular}
 	}
 	\caption{Precision of the top 100 retrieved documents on \emph{TMC} dataset. }
 	\label{tab:tmc}
 	\vspace{-4mm}
 \end{table}

Another observation is that the retrieval results tend to drop a bit when we set the length of hashing codes to be 64 or larger, which also happens for some baseline models.
This phenomenon has been reported previously in \citet{wang2012semi, liu2012supervised, wang2013semantic, chaidaroon2017variational}, and the reasons could be twofold: ($i$) for longer codes, the number of data points that are assigned to a certain binary code decreases exponentially.
As a result, many queries may fail to return any neighbor documents \citep{wang2012semi};
($ii$) considering the  size of training data, it is likely that the model may overfit with long hash codes \cite{chaidaroon2017variational}.
However, even with longer hashing codes, our NASH models perform stronger than the baselines in most cases (except for the 20Newsgroups dataset), suggesting that NASH can effectively allocate documents to informative/meaningful hashing codes even with limited training data.

We also evaluate the effectiveness of NASH in a \emph{supervised} scenario on the Reuters dataset, where the label or topic information is utilized during training.
As shown in  Figure~\ref{fig:sup}, our NASH-DN-S model consistently outperforms several supervised semantic hashing baselines, with various choices of hashing bits.
Notably, our model exhibits higher Top-100 retrieval precision than VDSH-S and VDSH-SP, proposed by \citet{chaidaroon2017variational}.
This may be attributed to the fact that in VDSH models, the continuous embeddings are not optimized with their future binarization in mind, and thus could hurt the relevance of learned binary codes.
On the contrary, our model is optimized in an end-to-end manner, where the gradients are directly backpropagated to the inference network (through the binary/discrete latent variable), and thus gives rise to a more robust hash function.

\subsection{Ablation study}\label{sec:ablation}
%

\subsubsection{The effect of stochastic sampling}
\begin{figure}[!h] 
	\centering
	\vspace{0mm}
	\includegraphics[width=.36\textwidth]{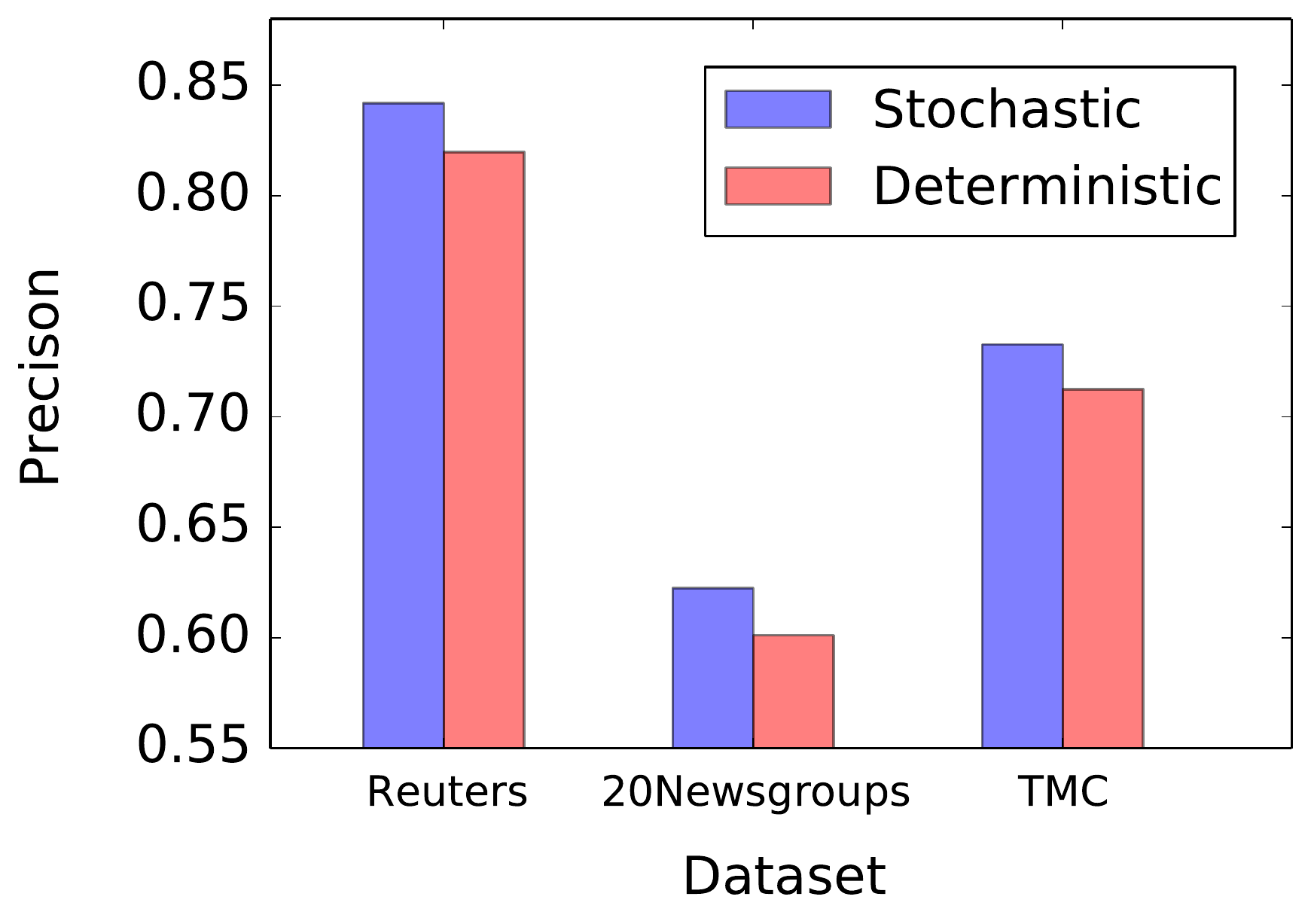}
	\caption{The precisions of the top 100 retrieved documents for NASH-DN with \textcolor{blue}{\emph{stochastic}} or \textcolor{red}{\emph{deterministic}} binary latent variables. }
	\label{fig:compare}
	\vspace{0mm}
\end{figure}

As described in Section~\ref{sec:method}, the binary latent variables $z$ in NASH can be either deterministically (via \eqref{eq:deter}) or stochastically (via \eqref{eq:stoc}) sampled.
We compare these two types of binarization functions in the case of unsupervised hashing.
As illustrated in Figure~\ref{fig:compare}, stochastic sampling shows stronger retrieval results on all three datasets, indicating that endowing the sampling process of latent variables with more stochasticity improves the learned representations.

\subsubsection{The effect of encoder/decoder networks}

\begin{table*}[ht] \small 
	\centering
	\begin{tabular}{c||c|c|c}
		\toprule[1.2pt]
		\textbf{Category} &  \textbf{Title/Subject} & \textbf{8-bit code}  & \textbf{16-bit code}  \\
		\hline
		\multirow{4}{*}{Baseball} & \emph{Dave Kingman for the hall of fame} & 1 1  1 0 1 0 0 1 & 0 0 1 0 1 1 0 1 0 0 0 0 0 1 1 0\\ 
		& \emph{Time of game} & 1 1  1 1 1 0 0 1 & 0 0 1 0 1 0 0 1 0 0 0 0 0 1 1 1\\ 
		& \emph{Game score report} & 1 1  1 0 1 0 0 1 & 0 0 1 0 1 1 0 1 0 0 0 0 0 1 1 0 \\
		& \emph{Why is Barry Bonds not batting 4th?} & 1 1  1 0 1 1 0 1 & 0 0 1 1 1 1 0 1 0 0 0 0 0 1 1 0 \\
		\hline
		\multirow{4}{*}{Electronics} & \emph{Building a UV flashlight} & 1 0 1 1 0 1 0 0 & 0 0 1 0 0 0 1 0 0 0 1 0 1 0 1 1 \\ 
		& \emph{How to drive an array of LEDs} & 1 0 1 1 0 1 0 1 & 0 0 1 0 0 0 1 0 0 0 1 0 1 0 0 1  \\ 
		& \emph{2\% silver solder} & 1 1 0 1 0 1 0 1 & 0 0 1 0 0 0 1 0 0 0 1 0 1 0 1 1  \\
		& \emph{Subliminal message flashing on TV} & 1 0 1 1 0 1 0 0 & 0 0 1 0 0 1 1 0 0 0 1 0 1 0 0 1 \\
		\bottomrule[1.2pt]
	\end{tabular}
	\caption{Examples of learned compact hashing codes on \emph{20Newsgroups} dataset. }
	\vspace{-1mm}
	\label{tab:case}
\end{table*}
Under the variational framework introduced here, the encoder network, \emph{i.e.}, hash function, and decoder network are jointly optimized to abstract semantic features from documents.
An interesting question concerns what types of network should be leveraged for each part of our NASH model.
In this regard, we further investigate the effect of using an encoder or decoder with different non-linearity, ranging from a linear transformation to two-layer MLPs.
We employ a base model with an encoder of two-layer MLPs and a linear decoder (the setup described in Section~\ref{sec:method}), and the ablation study results are shown in Table~\ref{tab:decoder}.

\begin{table} [ht!]
	\centering
	\def\arraystretch{1.2}
	\footnotesize 
	\begin{tabular}{c||c|c}
		\toprule[1.2pt]
		\textbf{Network} &  	\textbf{Encoder} & \textbf{Decoder}  \\
		\hline
		linear        & 0.5844 & 0.6225  \\
		one-layer MLP   & 0.6187 & 0.3559   \\ 
		two-layer MLP      & 0.6225 & 0.1047    \\
		\bottomrule[1.2pt]
	\end{tabular}
	\caption{Ablation study with different encoder/decoder networks. }
	\label{tab:decoder}
	\vspace{-1mm}
\end{table}

It is observed that for the encoder networks, increasing the non-linearity by stacking MLP layers leads to better empirical results.
In other words, endowing the hash function with more modeling capacity is advantageous to retrieval tasks.
However, when we employ a non-linear network for the decoder, the retrieval precision drops dramatically.
It is worth noting that the only difference between linear transformation and one-layer MLP is whether a non-linear activation function is employed or not. 

This observation may be attributed the fact that the decoder networks can be considered as a similarity measure between latent variable $z$ and the word embeddings $E_k$ for every word, and the probabilities for words that present in the document is maximized to ensure that $z$ is informative.
As a result, if we allow the decoder to be too expressive (\emph{e.g.}, a one-layer MLP),  it is likely that we will end up with a very flexible similarity measure but relatively less meaningful binary representations.
This finding is consistent with several image hashing methods, such as SGH \cite{dai2017stochastic} or binary autoencoder \cite{carreira2015hashing}, where a linear decoder is typically adopted to obtain promising retrieval results.
However, our experiments may not speak for other choices of encoder-decoder architectures, \emph{e.g.}, LSTM-based sequence-to-sequence models \cite{sutskever2014sequence} or DCNN-based autoencoder \cite{zhang2017deconvolutional}.

\subsection{Qualitative Analysis}
\subsubsection{Analysis of Semantic Information} \label{embeddings}

To understand what information has been learned in our NASH model, we examine the matrix $E \in \mathbb{R}^{d \times l}$ in \eqref{eq:decoder}.
Similar to \cite{miao2016neural, larochelle2012neural}, we select the 5 nearest words according to the word vectors learned from NASH and compare with the corresponding results from NVDM. 

As shown in Table~\ref{tab:words}, although our NASH model contains a binary latent variable, rather than a continuous one as in NVDM, it also effectively group semantically-similar words together in the learned vector space.
This further demonstrates that the proposed generative framework manages to bypass the binary/discrete constraint and is able to abstract useful semantic information from documents.

\subsubsection{Case Study}
In Table~\ref{tab:case}, we show some examples of the learned binary hashing codes on \emph{20Newsgroups} dataset.
We observe that for both 8-bit and 16-bit cases, NASH typically compresses documents with shared topics into very similar binary codes.
On the contrary, the hashing codes for documents with different topics exhibit much larger Hamming distance.
As a result, relevant documents can be efficiently retrieved by simply computing their Hamming distances. 

\section{Conclusions}
This paper presents a first step towards \emph{end-to-end} semantic hashing, where the binary/discrete constraints are carefully handled with an effective gradient estimator. A neural variational framework is introduced to train our model. Motivated by the connections between the proposed method and \emph{rate-distortion theory}, we inject data-dependent noise into the Bernoulli latent variable at the training stage. The effectiveness of our framework is demonstrated with extensive experiments. 

\paragraph{Acknowledgments} We would like to thank the ACL reviewers for their
insightful suggestions. This research was supported in part by DARPA, DOE, NIH, NSF and ONR.

\bibliography{acl2018}
\bibliographystyle{acl_natbib}
\end{document}